# On the optimality of tree-reweighted max-product message-passing

**Vladimir Kolmogorov**
Microsoft Research
Cambridge, UK

**Martin J. Wainwright**[*]
Departments of Statistics and EECS
UC Berkeley, Berkeley, CA

## Abstract

Tree-reweighted max-product (TRW) message passing [9] is a modified form of the ordinary max-product algorithm for attempting to find minimal energy configurations in Markov random field with cycles. For a TRW fixed point satisfying the strong tree agreement condition, the algorithm outputs a configuration that is provably optimal. In this paper, we focus on the case of binary variables with pairwise couplings, and establish stronger properties of TRW fixed points that satisfy only the milder condition of weak tree agreement (WTA). First, we demonstrate how it is possible to identify part of the optimal solution—i.e., a provably optimal solution for a subset of nodes— without knowing a complete solution. Second, we show that for submodular functions, a WTA fixed point always yields a globally optimal solution. We establish that for binary variables, any WTA fixed point always achieves the global maximum of the linear programming relaxation underlying the TRW method.

## 1 Introduction

Markov random fields (MRFs) provide a powerful framework for capturing dependencies among large collections of random variables [8]. One problem is determining a most probable configuration, or equivalently, a configuration with minimal energy. For tree-structured MRFs, minimal energy configurations can be computed efficiently by the max-product algorithm [8]. For graphs on cycles, in contrast, the problem of computing minimal energy configurations is intractable in the general setting, which motivates the development of approximate methods for attempting to find a minimum energy configuration. One widely-used heuristic is to apply the max-product algorithm to an MRF with cycles. This method, though convergent on trees, may fail to converge when applied to a graph with cycles. Moreover, although there are certain "local optimality" guarantees associated with any max-product fixed point [3, 10], it is also straightforward to construct problems for which the max-product algorithm will yield a non-optimal configuration.

In contrast to the ordinary max-product algorithm, the class of tree-reweighted (TRW) max-product algorithms [9] have certain correctness guarantees. In particular, for any TRW fixed point that satisfies *strong tree agreement*, the algorithm outputs a provably optimal configuration. However, not all fixed points satisfy this requirement; in general, they are characterized by a milder condition known as *weak tree agreement* (WTA) [6]. The main contribution of this paper is to establish a number of optimality properties of TRW fixed points that satisfy only the WTA condition. Our work focuses in particular on the case of binary variables with pairwise couplings. First, we establish that even when strong agreement does not hold, a WTA fixed point can still be used to provably certify a subset of an optimal configuration. Second, we prove that strong agreement can always be obtained for the class of submodular energy functions. Third, we show that for binary variables, a WTA fixed point always specifies a global optimum of the linear programming (LP) relaxation that underlies TRW message-passing.

The results and analysis of this paper focus primarily on the TRW message-passing updates. However, the link to the TRW-LP (namely, the proof that any WTA fixed point is a dual TRW-LP solution for binary problems) allows us to relate our results to other work on LP formulations. Although it is known that submodular problems can be solved by flow-based [1] and other LP formulations [2], our work establishes that distributed TRW message-passing also solves such problems. Other work [2, 5] has studied LP relaxations closely related to the TRW-LP, and established performance guarantees for certain MRFs (e.g., for metric labeling). The results of this paper differ qualitatively from this work: instead of bounding the quality of an approximate solution, we provide conditions for the TRW method to exactly specify (a subset of) variables in a globally optimal configuration. Our results also have connections with prior work on binary integer programming from the combinatorial optimization literature [1, 4]. In particular, previous work [4] has established a weak persistency property for the so-called Rhys relaxation. This result turns out to be related to our characterization of WTA fixed points, as we discuss at more length following Theorem 4.

[*]MW partially supported by an Intel Equipment grant and UC Berkeley junior faculty research grant.

The remainder of this paper is organized as follows. Section 2 is devoted to preliminaries, including notation and background on Markov random fields and max-product message-passing. In Section 3, we describe tree-reweighted (TRW) message-passing algorithms and their basic properties. Section 4 contains statements and proofs of our theoretical guarantees on TRW message-passing. We present some experimental results in Section 5, and conclude with a discussion in Section 6.

## 2 Preliminaries

We begin with necessary background on Markov random fields, the ordinary max-product algorithm, and tree-reweighted message-passing algorithms.

**Energy functions and MRFs:** Consider an undirected graph $G = (V, E)$, consisting of a set of vertices $V$ and a set of edges $E$. We denote by $(s, t)$ the edge between vertex $s$ and $t$ (or equivalently, for this undirected graph, between $t$ and $s$). A discrete Markov random field (MRF) is defined by associating to each vertex $s \in V$ a variable $x_s$ taking values in some discrete space $\mathcal{X}_s$. Of primary interest in this paper is the binary case $\mathcal{X}_s = \{0, 1\}$. By concatenating the variables at each node, we obtain a vector $\mathbf{x}$ with $n = |V|$ elements, taking values in $\mathcal{X}^n = \{0, 1\}^n$. Unless noted otherwise, we use symbols $s$ and $t$ to denote nodes in $V$, and $j$ and $k$ to denote particular values in $\{0, 1\}$.

In this paper, we consider Markov random fields of the form $p(\mathbf{x}; \theta) \propto \exp\{-E(\mathbf{x}; \theta)\}$, where $E$ is an *energy function*. We parameterize the energy function in terms of the following collection of functions. For each $s \in V$ and value $j \in \{0, 1\}$, let $\delta_j(x_s)$ denote an indicator function that is equal to one if $x_s = j$, and zero otherwise. Taking products of the form $\delta_j(x_s) \delta_k(x_s)$ yields indicator functions for the event $\{x_s = j, x_t = k\}$. Lastly, we define a constant function $\phi_{\text{const}}(\mathbf{x}) = 1$ for all $\mathbf{x} \in \mathcal{X}^n$. With this set-up, the *canonical overcomplete representation* is given by

$$\{\phi_{\text{const}}(\mathbf{x})\} \cup \{\delta_j(x_s) \mid j \in \{0,1\}, s \in V\} \cup$$
$$\{\delta_j(x_s)\delta_k(x_t) \mid (j, k) \in \{0,1\}^2, (s,t) \in E\}. \quad (1)$$

Note that there holds $(st; jk) \equiv (ts; kj)$, so that $\theta_{st;jk}$ and $\theta_{ts;kj}$ are the same element. It will sometimes be convenient to denote elements $\theta_{s;j}$ and $\theta_{st;jk}$ by $\theta_s(j)$ and $\theta_{st}(j, k)$, respectively. Finally, we use the overcomplete potentials (1) and parameters $\theta$ to define single node functions $\theta_s(x_s) = \sum_j \theta_{s;j}\delta_j(x_s)$ and edgewise functions $\theta_{st}(x_s, x_t) = \sum_{j,k} \theta_{st;jk}\delta_j(x_s)\delta_k(x_t)$.

With these notational conventions, the energy function $E(\mathbf{x}; \theta)$ can be decomposed as a constant term plus a sum over the vertices and edges of the graph in the following way:

$$E(\mathbf{x}; \theta) = \theta_{\text{const}} + \sum_{s \in V} \theta_s(x_s) + \sum_{(s,t) \in E} \theta_{st}(x_s, x_t). \quad (2)$$

An important property of the representation (1) is its *overcompleteness*, meaning that many different parameter vectors $\theta$ can be used to parameterize the same energy function. If two parameter vectors $\theta$ and $\theta'$ define the same energy function (i.e. $E(\mathbf{x}; \theta') = E(\mathbf{x}; \theta)$ for all $\mathbf{x} \in \mathcal{X}^n$), then $\theta'$ is called a *reparameterization* of $\theta$, and the relation is denoted by $\theta' \cong \theta$. As a particular example, it can be seen the energy function is preserved by adding a constant to $\theta_{\text{const}}$, and then subtracting the same constant from the vector $\theta_1 = \{\theta_{1;j}, j \in \mathcal{X}_1\}$, so that the underlying parameter vectors represent a reparameterization.

**Min-marginals:** A min-marginal is a function that provides information about the minimum values of the energy under different constraints. In precise terms, we define $\Phi(\theta) := \min_{\mathbf{x} \in \mathcal{X}^n} E(\mathbf{x}; \theta)$, and

$$\Phi_{s;j}(\theta) := \min_{\{\mathbf{x} \in \mathcal{X}^n \mid x_s = j\}} E(\mathbf{x}; \theta) \quad (3a)$$
$$\Phi_{st;jk}(\theta) := \min_{\{\mathbf{x} \in \mathcal{X}^n \mid x_s = j, \, x_t = k\}} E(\mathbf{x}; \theta). \quad (3b)$$

We refer to the functions $\Phi_{s;j}(\theta)$ and $\Phi_{st;jk}(\theta)$ as the *min-marginals* for node $s$ and edge $(s, t)$ respectively.

**Max-product and normal form:** The max-product (min-sum) algorithm [8] is used for exact computation of minimal energy configurations in trees, as well as for approximate computation in graphs with cycles. Although max-product is usually specified in terms of message-passing, it can also be formulated in terms of reparameterization operations on the vector $\theta$. More concretely, sending a message from node $s$ to node $t$ is equivalent to performing a certain reparameterization of vectors $\theta_{st}$ and $\theta_t$. In this context, it can be shown [10] that the vector $\theta$ is a min-sum fixed point if and only if it satisfies the following conditions for each direction $(s \to t)$ of every edge in the graph:

$$\min_{j \in \mathcal{X}_s} \{\theta_{s;j} + \theta_{st;jk}\} = \text{const}_{st} \quad \forall\, k \in \mathcal{X}_t, \quad (4)$$

where $\text{const}_{st}$ is a constant independent of $j$ and $k$. With this set-up, we say that a vector $\theta$ is in *normal form* if it satisfies equation (4) for each direction of every edge in the graph. We say that $\theta$ is in *canonical normal form* if in addition it satisfies the following conditions:

$$\min_{j \in \mathcal{X}_s} \theta_{s;j} = 0 \quad \forall\, s \in V \quad (5a)$$

$$\min_{(j,k)} \left(\theta_{s;j} + \theta_{st;jk} + \theta_{t;k}\right) = 0 \quad \forall\, (s,t) \in E. \quad (5b)$$

Any vector $\theta$ in a normal form can be reparameterized into a canonical normal form by subtracting a constant from vectors $\theta_s$ and $\theta_{st}$ and adding the same constant to $\theta_{\text{const}}$. Moreover, it can be seen that whenever $\theta$ is in canonical normal form, then the constant $\text{const}_{st}$ in equation (4) must be zero (see [6]). If graph $G$ is a tree and $\theta$ is in canonical form, then the values $\theta_{s;j}$ and $\theta_{s;j} + \theta_{st;jk} + \theta_{t;k}$ are equivalent [10], up to an additive constant offset, to the min-marginals defined in equation (3), and there holds $\Phi(\theta) = \theta_{\text{const}}$.

## 3 Tree-reweighted message-passing

At a high level, the underlying motivation of tree-reweighted message-passing [9] is to maximize a lower bound on the the minimal energy based on a convex combination of min-marginal values defined by trees. As we describe here, this lower bound is tight when the strong tree agreement condition holds, in which case the algorithm outputs a minimum energy configuration. Weak tree agreement (WTA) is a milder condition that is satisfied by all TRW fixed points.

**Convex combinations of trees:** Let $\mathfrak{T}$ be a collection of trees contained in the graph $G$, and let $\rho := \{\rho(T) \,|\, T \in \mathfrak{T}\}$ be a probability distribution on $\mathfrak{T}$. Throughout the paper, we assume that each tree $T \in \mathfrak{T}$ has a non-zero probability (i.e. $\rho(T) > 0$), and that each edge in graph $G$ is covered by at least one tree. For a given tree $T = (V(T), E(T))$, we define the subset $\mathcal{I}(T) \subset \mathcal{I}$ as follows

$$\{\text{const}\} \cup \{(s; j) \,|\, s \in V(T)\} \cup \{(st; jk) \,|\, (s,t) \in E(T)\},$$

corresponding to those indexes associated with vertices and edges in the tree.

Given a tree $T \in \mathfrak{T}$, an energy parameter $\theta(T)$ is *tree-structured* if it belongs to the constraint set

$$\mathcal{A}(T) := \{\theta(T) \in \mathbb{R}^d \,|\, \theta_\alpha(T) = 0 \;\; \forall\, \alpha \in \mathcal{I}(T) \backslash \mathcal{I}\}. \quad (6)$$

In words, any member of $\mathcal{A}(T)$ is a vector of length $|\mathcal{I}|$ with zeros in all elements *not* corresponding to vertices or edges in the tree. Concatenating a set of tree-structured vectors (one for each tree $T \in \mathfrak{T}$) yields a larger vector $\vec{\theta} = \{\theta(T) \,|\, T \in \mathfrak{T}\}$, which by definition is a member of the set $\mathcal{A} := \{\vec{\theta} \in \mathbb{R}^{d \times |T|} \,|\, \theta(T) \in \mathcal{A}(T) \text{ for all } T \in \mathfrak{T}\}$.

For each tree $T$, we consider the associated min-function $\Phi_T(\theta(T)) = \min_{\mathbf{x} \in \mathcal{X}^n} E(\mathbf{x}; \theta(T))$. As the minimum of a set of linear functions, each such function $\Phi_T$ is concave in $\theta$. We now define a new function $\Phi_\rho : \mathcal{A} \to \mathbb{R}$ as a the convex combination $\Phi_\rho(\vec{\theta}) := \sum_{T \in \mathfrak{T}} \rho(T) \Phi_T(\theta(T))$. Since each $\Phi_T$ is concave, the function $\Phi_\rho$ is also concave. It can be shown [9, 6] that for any vector $\vec{\theta} \in \mathcal{A}$ that satisfies the relation $\sum_{T \in \mathfrak{T}} \rho(T) \theta(T) \cong \bar{\theta}$, the value $\Phi_\rho(\vec{\theta})$ is a lower bound on the optimal value of the energy for vector $\bar{\theta}$. This lower bound is the motivation for considering the following constrained maximization problem:

$$\max_{\vec{\theta} \in \mathcal{A}} \Phi_\rho(\vec{\theta}) \quad \text{subject to} \quad \sum_{T \in \mathfrak{T}} \rho(T) \theta(T) \cong \bar{\theta}. \quad (7)$$

Note that the constraint means that the parameters $\sum_{T \in \mathfrak{T}} \rho(T) \theta(T)$ and $\bar{\theta}$ are different parameterizations of the same energy function. This constraint is equivalent to a set of linear constraints on $\vec{\theta}$.

**Tree agreement:** A number of different *tree-reweighted message-passing* (hereafter TRW) algorithms have been developed [9, 6]. They share the common goal of maximizing the function $\Phi_\rho$, and all maintain the constraint in equation (7). The sequential TRW algorithm [6] also has the desirable property that of never decreasing the function $\Phi_\rho$, and is guaranteed to have a limit point that satisfies the *weak tree agreement* (WTA) condition.

For any energy parameter vector $\theta$, let $\text{OPT}(\theta) := \{\mathbf{x} \in \mathcal{X}^n \,|\, E(\mathbf{x}; \theta) = \Phi(\theta)\}$ be the set of configurations $\mathbf{x} \in \mathcal{X}^n$ that are optimal for the energy function defined by $\theta$. Note that for each $\theta$, $\text{OPT}(\theta)$ is a particular subset of $\mathcal{X}^n$. Naturally, the notation $\text{OPT}(\theta(T))$ denotes the set of optimal configurations for the tree-structured energy function defined by $\theta(T)$. Given a collection $\vec{\theta} \in \mathcal{A}$ of tree-structured parameters, we define the Cartesian product $\text{OPT}(\vec{\theta}) := \bigotimes_{T \in \mathfrak{T}} \text{OPT}(\theta(T))$ of all the sets of optimal configurations for vectors $\theta(T)$, as $T$ ranges over $\mathfrak{T}$.

We say that $\vec{\theta}$ satisfies the *tree agreement* (TA) condition if the intersection $\cap_{T \in \mathfrak{T}} \text{OPT}(\theta(T))$ is non-empty, meaning that there exists at least one configuration $\mathbf{x}^*$ that is optimal for each one of the trees. The significance of tree agreement is demonstrated by the following result:

**Theorem 1.** *[9] For some given $\rho$, suppose that the vector $\vec{\theta}$ obeys the constraint in equation (7), and satisfies the TA condition with configuration $\mathbf{x}^*$. Then the configuration $\mathbf{x}^*$ is minimal energy for $E(\mathbf{x}; \bar{\theta})$.*

Therefore, the strong TA condition dictates whether a given vector $\vec{\theta}$ can be used to find an optimal configuration.

**Weak tree agreement:** For analyzing fixed points of tree-reweighted message-passing, it turns out to be useful to introduce a refined notion, to which we refer as the *weak tree agreement* (WTA) condition. In a certain sense, this condition characterizes local maxima of the algorithm with respect to function $\Phi_\rho$. More precisely, once WTA condition has been achieved, the value of function $\Phi_\rho$ will not change [6].

For each tree $T$, let $\mathbb{S}(T) \subseteq \mathcal{X}^n$ be a set of configurations, and let $\mathbb{S} = \otimes_{T \in \mathfrak{T}} \mathbb{S}(T)$ be the Cartesian product of all these subsets. We say that the family $\mathbb{S}$ is *consistent* if it satisfies the following three conditions:
  (a) For any tree $T$, the set $\mathbb{S}(T)$ is non-empty.
  (b) Let $s$ be a vertex belonging to both trees $T$ and $T'$. Then for any configuration $\mathbf{x} \in \mathbb{S}(T)$, there exists a configuration $\mathbf{x}' \in \mathbb{S}(T')$ such that $x_s = x'_s$.
  (c) Suppose that edge $(s,t)$ is contained in trees $T$ and $T'$. Then for any configuration $\mathbf{x} \in \mathbb{S}(T)$, there exists a configuration $\mathbf{x}' \in \mathbb{S}(T')$ such that $(x_s, x_t) = (x'_s, x'_t)$.

We then say that the vector $\vec{\theta} = \{\theta(T) \mid T \in \mathfrak{T}\}$ satisfies the *weak tree agreement* condition if there exists a family $\mathbb{S} \subseteq \mathrm{OPT}(\vec{\theta})$ that is consistent. Note that the WTA condition is a generalization of the TA condition; more precisely, it is easy to see that the TA condition implies the WTA condition.

## 4 Optimality properties

The analysis in the remainder of the paper is made under the following standing assumptions:
**A1.** The variable spaces are binary at each node (i.e., $\mathcal{X}_s = \{0,1\}$ for all $s \in V$).
**A2.** We have found a collection $\vec{\theta} = \{\theta(T) \mid T \in \mathfrak{T}\}$ such that (i) It satisfies the WTA condition, and (ii) it specifies a $\rho$-parameterization of $\bar{\theta}$, meaning that $\sum_{T \in \mathfrak{T}} \rho(T)\theta(T) \cong \bar{\theta}$.

It can be shown [9] that there always exists a vector $\vec{\theta}$ that satisfies A2. Moreover, Theorem 1 guarantees that a minimum energy configuration can be obtained whenever $\vec{\theta}$ satisfies (strong) tree agreement. Our goal is analyze properties of $\vec{\theta}$ when it only satisfies *weak* tree agreement.

The following concepts are useful in our analysis. Given a consistent family $\mathbb{S}$, we define for each node $s \in V$ the set $\chi_s(\mathbb{S}) \subseteq \{0,1\}$ of possible values of $x_s$ for configurations $\mathbf{x} \in \mathbb{S}(T)$, where tree $T \in \mathfrak{T}$ contains node $s$. Since $\mathbb{S}$ is a consistent family, this definition does not depend on the tree (as long as it contains $s$). Moreover, for an edge $(s,t)$, we define the set $\chi_{st}(\mathbb{S}) \subseteq \{0,1\} \times \{0,1\}$ in an analogous manner. More precisely, we define $\chi_s(\mathbb{S}) := \{j \in \{0,1\} \mid \exists T \in \mathfrak{T}_s, \mathbf{x} \in \mathbb{S}(T) \text{ s.t. } x_s = j\}$; similarly, the set $\chi_{st}(\mathbb{S})$ is given by

$$\{(j,k) \mid \exists T \in \mathfrak{T}_{st}, \mathbf{x} \in \mathbb{S}(T) \text{ s.t. } (x_s, x_t) = (j,k)\}.$$

where $\mathfrak{T}_s$ and $\mathfrak{T}_{st}$ denote the sets of trees in $\mathfrak{T}$ containing node $s$ and edge $(s,t)$, respectively. The following properties are easy to verify:

**Lemma 1.** *For any consistent family $\mathbb{S}$, the optimal local sets satisfy the following properties:*

*(a) If $(j,k) \in \chi_{st}(\mathbb{S})$, then $j \in \chi_s(\mathbb{S})$ and $k \in \chi_t(\mathbb{S})$.*

*(b) If $j \in \chi_s(\mathbb{S})$ then there exists $k \in \mathcal{X}_t$ such that $(j,k) \in \chi_{st}(\mathbb{S})$.*

*(c) For any vector $\gamma$ in canonical normal form:*
*(i) If $j \in \chi_s(\mathbb{S})$, then $\gamma_s(j) = 0$; and*
*(ii) If $(j,k) \in \chi_{st}(\mathbb{S})$, then $\gamma_{st}(j,k) = 0$.*

**Correctness of individual variables;** In this section, we show that even if the strong TA condition is not satisfied, a WTA fixed point $\vec{\theta}$ can still yield useful information about a subset of an optimal solution.

**Theorem 2.** *Let $\vec{\theta}$ be a WTA fixed point, with corresponding consistent collection $\mathbb{S} \subseteq OPT(\vec{\theta})$. Let $V^{\mathrm{fix}}$ be the set of vertices such that $\chi_s(\mathbb{S})$ contains a single element $j_s$. Then it is always possible to find a minimum energy configuration $\mathbf{x}^*$ of $E(\mathbf{x}; \bar{\theta})$ such that $x_s^* = j_s$ for all $s \in V^{\mathrm{fix}}$.*

*Proof.* To start, we assume without loss of generality that for every tree $T \in \mathfrak{T}$, the parameter $\theta(T)$ is in canonical normal form and $\theta(T)_{\mathrm{const}} = 0$. (If not, it can be converted to this form running the ordinary max-product algorithm for the tree; the key properties of the collection $\mathrm{OPT}(\vec{\theta})$ are not modified. Moreover, the assumption about the constant term does not affect the theorem). We then define $V^{\mathrm{fr}} := V \setminus V^{\mathrm{fix}}$, corresponding to the set of "free" vertices (i.e., not fixed). The two subsets $V^{\mathrm{fr}}$ and $V^{\mathrm{fix}}$ induce particular subgraphs of the original graph, which we denote by $G^{\mathrm{fr}} = (V^{\mathrm{fr}}, E^{\mathrm{fr}})$ and $G^{\mathrm{fix}} = (V^{\mathrm{fix}}, E^{\mathrm{fix}})$ respectively. As illustrated in Figure 1, these subgraphs induce partitions of the vertex and edge set, where $E^{\mathrm{bou}} = \{(s,t) \in E \mid s \in V^{\mathrm{fr}}, t \in V^{\mathrm{fix}}\}$ is the set of boundary edges crossing between $V^{\mathrm{fix}}$ and $V^{\mathrm{fr}}$. This graph-based decomposition induces a parallel decomposition of the vector $\widehat{\theta} := \sum_{T \in \mathfrak{T}} \rho(T)\theta(T)$ into three parts $\widehat{\theta}^{\text{fix}} = \widehat{\theta}(G^{\text{fix}})$, $\widehat{\theta}^{\text{fr}} = \widehat{\theta}(G^{\text{fr}})$, and $\widehat{\theta}^{\text{bou}} = \widehat{\theta}(E^{\text{bou}})$ as follows

$$\widehat{\theta} = \widehat{\theta}^{\text{fix}} + \widehat{\theta}^{\text{fr}} + \widehat{\theta}^{\text{bou}} \tag{8a}$$
$$E(\mathbf{x}\,;\widehat{\theta}) = E(\mathbf{x}\,;\widehat{\theta}^{\text{fix}}) + E(\mathbf{x}\,;\widehat{\theta}^{\text{fr}}) + E(\mathbf{x}\,;\widehat{\theta}^{\text{bou}}). \tag{8b}$$

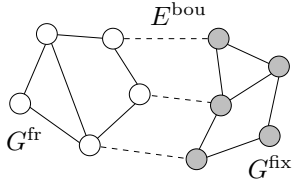

**Figure 1.** Partition of the graph $G$ into a subgraph $G^{\text{fix}} = G[V^{\text{fix}}]$ induced by the fixed vertices, and a subgraph $G^{\text{fr}} = G[V^{\text{fr}}]$ induced by free vertices. The boundary edges $E^{\text{bou}}$ cross between these two node-induced subgraphs.

With this set-up, our approach is to define a configuration $\mathbf{x}^*$, and then prove that it is minimum energy for all three components in the decomposition (8b). Doing so guarantees that $\mathbf{x}^*$ is a minimum energy configuration for $E(\mathbf{x}\,;\bar{\theta})$, since

$$E(\mathbf{x}^*\,;\bar{\theta}) \stackrel{(a)}{=} E(\mathbf{x}^*\,;\widehat{\theta})$$
$$\stackrel{(b)}{\leq} E(\mathbf{x}\,;\widehat{\theta}^{\text{fix}}) + E(\mathbf{x}\,;\widehat{\theta}^{\text{fr}}) + E(\mathbf{x}\,;\widehat{\theta}^{\text{bou}})$$
$$\stackrel{(c)}{=} E(\mathbf{x}\,;\bar{\theta}).$$

where equalities (a) and (c) follow since $\bar{\theta} \cong \widehat{\theta}$; and inequality (b) follows from decomposition (8b) and the optimality of $\mathbf{x}^*$ is for each of the three components.

We construct $\mathbf{x}^*$ in the following way. First, for all $s \in V^{\text{fix}}$, define $x_s^* = j_s$. Second, for nodes $s \in V^{\text{fr}}$, choose some $\mathbf{z}^* \in \arg\min_{\mathbf{x}} E(\mathbf{x}\,;\widehat{\theta}(G^{\text{fr}}))$, and set $x_s^* = z_s^*$. By definition, these choices ensure that $\mathbf{x}^*$ is a minimum energy configuration for $E(\mathbf{x}\,;\widehat{\theta}(G^{\text{fr}}))$. Next we prove that the same holds for the other two components:

**Component $\widehat{\theta}(G^{\text{fix}})$** Let us define a projection operation on $G^{\text{fix}} = (V^{\text{fix}}, E^{\text{fix}})$ as follows: given a vector $\theta$, we define

$$\Pi^{\text{fix}}(\theta)_\alpha = \begin{cases} \theta_\alpha & \text{if } \alpha \in \mathcal{I}(V^{\text{fix}}) \cup \mathcal{I}(E^{\text{fix}}) \\ 0 & \text{otherwise.} \end{cases} \tag{10}$$

Consider the tree-structured parameters $\theta(T)$ that form the collection $\vec{\theta}$; of interest to us are their projections $\gamma^T := \Pi^{\text{fix}}(\theta(T))$ onto the graph $G^{\text{fix}}$. Using definition (10) and Assumption A2, we have the decomposition $\widehat{\theta}(G^{\text{fix}}) \cong \sum_{T\in\mathfrak{T}} \rho(T)\gamma^T$, and we are guaranteed that each $\gamma^T$ is in canonical normal form. Moreover, by construction of $\mathbf{x}^*$, we have $\gamma_s^T(x_s^*) = 0$ for any node $s$ and $\gamma_{st}^T(x_s^*, x_t^*) = 0$ for any edge $(s,t)$. These properties guarantee that $E(\mathbf{x}^*\,;\gamma^T) = 0 = \gamma_{\text{const}}^T = \Phi_T(\gamma^T)$, implying that $\mathbf{x}^*$ is an optimal configuration for parameter vector $\gamma^T$. Consequently, the (strong) tree agreement condition is satisfied for $\widehat{\theta}(G^{\text{fix}})$ and the decomposition $\{\gamma^T \mid T \in \mathfrak{T}\}$; by Theorem 1, the vector $\mathbf{x}^*$ is optimal for component $\widehat{\theta}(G^{\text{fix}})$.

**Component $\widehat{\theta}(E^{\text{bou}})$** This case is slightly more complicated than the previous one, and requires the assumption that $\mathcal{X}_s = \{0,1\}$ for all $s \in V$ (which we have not yet used). Consider a boundary edge $(s,t) \in E^{\text{bou}}$ such that $s \in V^{\text{fix}}$ and $t \in V^{\text{fr}}$. By definition of $\mathbb{S}$ and $V^{\text{fix}}$, we have $\chi_s(\mathbb{S}) = \{j_s\}$ and $\chi_t(\mathbb{S}) = \{0,1\}$, from which Lemma 1 implies that $\chi_{st}(\mathbb{S}) = \{(j_s,0),(j_s,1)\}$. For some tree $T \in \mathfrak{T}$ containing edge $(s,t)$, let us define the shorthand $\gamma := \theta(T)$; of interest to us are the parameters $\gamma_s, \gamma_t$ and $\gamma_{st}$.

Since $\gamma = \theta(T)$ is in the canonical normal form, it must satisfy the fixed point condition (4) for each direction of every edge $(s,t) \in E(T)$. Moreover, Lemma 1 implies that

$$\gamma_s(j_s) = \gamma_t(0) = \gamma_t(1) = 0 \quad \text{and} \quad \gamma_s(1-j_s) \geq 0. \tag{11}$$

Therefore, the fixed point condition (4) for directed edge $(t \to s)$ reduces to

$$\min_{k\in\{0,1\}} (\gamma_{st}(j,k)) = 0 \quad \forall\, j \in \{0,1\}. \tag{12}$$

Moreover, the fact that $(j_s,0)$ and $(j_s,1)$ are both in $\chi_{st}(\mathbb{S})$ implies that $\Phi_{st;j_s0}(\gamma) = \Phi_{st;j_s1}(\gamma)$, which in turn implies that

$$\gamma_s(j_s) + \gamma_{st}(j_s,0) + \gamma_t(0) = \gamma_s(j_s) + \gamma_{st}(j_s,1) + \gamma_t(1).$$

This equation, in conjunction with equation (11), implies that $\gamma_{st}(j_s,0) = \gamma_{st}(j_s,1)$. Combining this equality with condition (12) for $j = j_s$ yields

$$\gamma_{st}(j_s,0) = \gamma_{st}(j_s,1) = 0. \tag{13}$$

We have shown that equation (13) holds for $\gamma = \theta(T)$ defined by any tree $T$ containing a boundary edge $(s,t)$. Since equation (13) also holds trivially for any tree $T$ *not* containing $(s,t)$, the same statement holds for the convex combination, $\widehat{\theta} = \sum_{T\in\mathfrak{T}} \rho(T)\theta(T)$. More specifically, for any boundary edge $(s,t) \in E^{\text{bou}}$, we have $\widehat{\theta}_{st}(j_s,0) = \widehat{\theta}_{st}(j_s,1) = 0$ and $\widehat{\theta}_{st}(1-j_s,0) \geq 0$, and $\widehat{\theta}_{st}(1-j_s,1) \geq 0$. These statements imply that $E(\mathbf{x}^*\,;\widehat{\theta}(E^{\text{bou}})) = 0$ and $E(\mathbf{x}\,;\widehat{\theta}(E^{\text{bou}})) \geq 0$ for all $\mathbf{x} \in \mathcal{X}^n$, meaning that $\mathbf{x}^*$ is an optimal configuration for vector $\widehat{\theta}(E^{\text{bou}})$ as claimed. □

An immediate consequence of Theorem 2 is that for any vertex $s \in V^{\text{fix}}$ with $\chi_s(\mathbb{S}) = \{j\}$, it necessarily holds that $\Phi_{s;1-j}(\bar{\theta}) \geq \Phi_{s;j}(\bar{\theta})$. In fact, this obvious result can be strengthened in the following way.

**Corollary 1.** *Assume without loss of generality that each vector $\theta(T)$ is in canonical normal form. Consider any node $s \in V^{\text{fix}}$ such that $\chi_s(\mathbb{S}) = \{j\}$. Then the quantity $C := \widehat{\theta}_{s;\bar{j}} = \sum_{T \in \mathfrak{T}} \rho(T) \theta_{s;\bar{j}}(T)$, where $\bar{j} := 1 - j$, gives a lower bound on the difference in min-marginals as follows:*

$$\Phi_{s;\bar{j}}(\bar{\theta}) \geq \Phi_{s;j}(\bar{\theta}) + C = \Phi(\bar{\theta}) + C. \qquad (14)$$

The proof is omitted due to space constraints (but the argument is similar to the previous theorem). An important consequence of Corollary 1 is that if $\theta_{s;\bar{j}}(T) > 0$ for at least one tree $T$, then it follows that $x_s^* = j$ in *all optimal configurations*.

**Optimality for submodular functions:** In this section, we prove that tree-reweighted message-passing is always guaranteed to compute a minimum energy configuration for *submodular* energy functions. An energy function is submodular if for every edge $(s,t) \in E$, the vector $\bar{\theta}_{st}$ satisfies the following inequality:

$$\bar{\theta}_{st;00} + \bar{\theta}_{st;11} \leq \bar{\theta}_{st;01} + \bar{\theta}_{st;10}. \qquad (15)$$

It is well-known that any submodular energy function can be minimized in polynomial time by reducing the problem to a maximum flow problem on an auxiliary graph [1]. The main result of this section is to show that an optimal solution can also be obtained by TRW message-passing, which suggests that TRW should also behave well for near-submodular functions (see Section 5). It should also be noted that the ordinary max-product algorithm does not have a similar guarantee, in that it may output a non-optimal configuration even for a submodular problem.

**Theorem 3.** *Suppose that assumptions A1 and A2 hold, $\vec{\theta}$ satisfies the WTA condition, and that the energy function $E(\mathbf{x}; \bar{\theta})$ is submodular. Using the notation of Theorem 2, consider configurations $\mathbf{x}$ and $\mathbf{y}$ defined as follows:*

$$x_s := \begin{cases} j_s & \text{if } s \in V^{\text{fix}} \\ 0 & \text{if } s \in V^{\text{fr}} \end{cases} \qquad y_s := \begin{cases} j_s & \text{if } s \in V^{\text{fix}} \\ 1 & \text{if } s \in V^{\text{fr}} \end{cases} \qquad (16)$$

*Then both $\mathbf{x}$ and $\mathbf{y}$ are optimal configurations for $\bar{\theta}$.*

*Proof.* Using the same notation as in the proof of Theorem 2, consider an edge $(s,t) \in E^{\text{fr}}$, and let us analyze the parameter $\theta(T)$ for some $T$ containing edge $(s,t)$. Since $s, t \in V^{\text{fr}}$ we have $\chi_s(\mathbb{S}) = \chi_t(\mathbb{S}) = \{0,1\}$; using Lemma 1, this implies that

$$\theta_{s;0}(T) = \theta_{s;1}(T) = \theta_{t;0}(T) = \theta_{t;1}(T) = 0. \quad (17)$$

Using equation (4), we conclude that all elements of vector $\gamma_{st}$ are non-negative.

We now claim that $\theta_{st;00}(T) = 0$ *for all* trees that contain edge $(s,t)$. Suppose that this were not the case—namely, that there exists some tree $T'$ such that the vector $\theta(T')$ satisfies $\theta_{st;00}(T') > 0$, whence $(0,0) \notin \chi_{st}(\mathbb{S})$. Using Lemma 1, this fact implies that $\{(0,1),(1,0)\} \subseteq \chi_{st}(\mathbb{S})$. Consequently, for all trees $T$ containing edge $(s,t)$, there must hold

$$\theta_{st;01}(T) = \theta_{st;10}(T) = 0. \qquad (18)$$

Since these properties must also hold for the convex combination $\widehat{\theta} := \sum_T \rho(T) \theta(T)$, we have thus established that

$$\widehat{\theta}_{st;00} > 0, \ \widehat{\theta}_{st;01} = \widehat{\theta}_{st;10} = 0, \ \text{and } \widehat{\theta}_{st;11} \geq 0. \qquad (19)$$

For binary problems, it can be shown [6] that the quantity $\widehat{\theta}_{st}(0,1) + \widehat{\theta}_{st}(1,0) - \widehat{\theta}_{st}(0,0) - \widehat{\theta}_{st}(1,1)$ is an invariant, equal to the same constant for any reparameterization of $\bar{\theta}$. Therefore, we have shown that $\widehat{\theta}_{st;01} + \widehat{\theta}_{st;10} - \widehat{\theta}_{st;00} - \widehat{\theta}_{st;11}$ is equal to $\bar{\theta}_{st;01} + \bar{\theta}_{st;10} - \bar{\theta}_{st;00} - \bar{\theta}_{st;11}$. Note that the first sum (involving $\widehat{\theta}$) is negative because of conditions (19), whereas the second sum (involving $\bar{\theta}$) is non-negative since $E(\mathbf{x}; \bar{\theta})$ is submodular. This contradiction establishes our claim.

Thus, we have established that for all edges $(s,t) \in E^{\text{fr}}$ and trees $T$, the vector $\theta(T)_{st}$ is non-negative with its $(0,0)$ element equal to zero. Combining this fact with the properties given in the proof of theorem 2 yields that for any tree $T \in \mathfrak{T}$, we have $\theta(T)_s(x_s) = 0$ for any node $s$, and and $\theta(T)_{st}(x_s, x_t) = 0$ for any edge $(s,t)$. Therefore, it follows that $E(\mathbf{x}; \theta(T)) = \theta(T)_{\text{const}} = \Phi(\theta(T))$, meaning that $\mathbf{x}$ is an optimal configuration for vector $\theta(T)$. Since this holds for every $T$, the family $\vec{\theta}$ satisfies the TA condition for $\mathbf{x}$, whence Theorem 1 yields that $\mathbf{x}$ is an optimal configuration for vector $\bar{\theta}$. A similar argument establishes that $\mathbf{y}$ is also an optimal configuration. □

**Global maximum of the lower bound** As discussed in Section 2, the underlying goal of TRW algorithms is to solve maximization problem (7), which can be reformulated as a linear program. Since this is a simple convex problem, one might expect that a TRW fixed point always specifies a global maximum of the lower bound (7). However, Kolmogorov [6] provided a counterexample to show that this is not the case in general. This counterexample involves an energy function with ternary state spaces. Interestingly, this fact turns out to crucial: as we now show, for functions of binary variables the two conditions are equivalent.

**Theorem 4.** *For binary problems, any vector $\vec{\theta}$ satisfying conditions A1 and A2 is a global maximum of problem (7).*

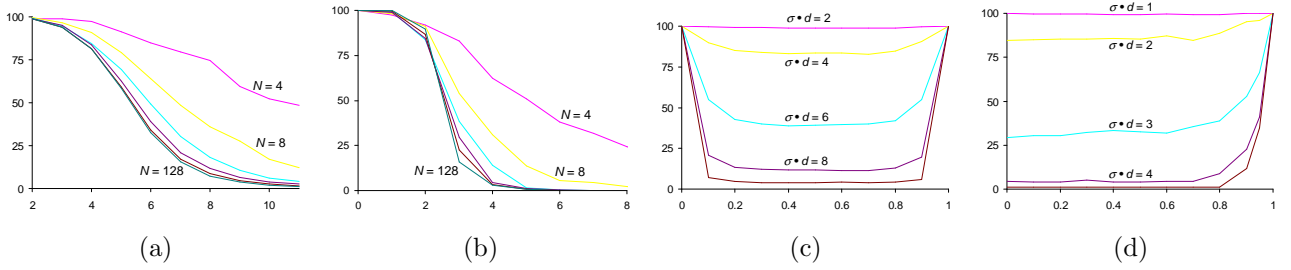

**Figure 2.** Panels (a) and (b): Plots of correct percentage $p_{\text{cor}}$ versus the potential strength $\sigma \cdot d$ (where $\sigma$ is interaction strength and $d$ is node degree) for grids (panel (a)) and complete graphs (panel (b)). Different curves in each panel correspond to different graph sizes $N \in \{4, 8, 16, 32, 64, 128\}$ (from top to bottom). Panels (c) and (d): Plots of correct percentage $p_{\text{cor}}$ versus the mixing parameter $\alpha \in [0, 1]$. for grids (panel (c)) complete graphs (panel (d)). Different curves in each panel correspond to different potential strengths $\sigma \cdot d \in \{2, 4, 6, 8, 10\}$ for grids and $\sigma \cdot d \in \{1, 2, 3, 4, 5\}$ for complete graphs (from top to bottom).

*Proof.* The proof is based on the Lagrangian dual associated with problem (7) (see [9]). More concretely, our approach is to specify a dual feasible solution, and then show that the associated dual value attains the primal value $\Phi(\vec{\theta})$, which guarantees optimality.

We assume without loss of generality that each vector $\theta(T)$ is in the canonical normal form. For any vector in this form [6], we have $\Phi_T(\theta(T)) = \theta(T)_{\text{const}}$, so that $\Phi_\rho(\vec{\theta}) = \widehat{\theta}_{\text{const}}$ where $\widehat{\theta} = \sum_{T \in \mathfrak{T}} \rho(T)\theta(T)$. The dual problem given in [9] is the linear program

$$\min_{\tau \in \mathbb{R}^d} \langle \bar{\theta}, \tau \rangle \;=\; \min_{\tau \in \mathbb{R}^d} \sum_{\alpha \in \mathcal{I}} \bar{\theta}_\alpha \tau_\alpha \qquad (20)$$

subject to $\tau$ belonging to the constraint set $\text{LOCAL}(G)$ defined by the constraints $\tau \in \mathbb{R}_+^d$; $\tau_{\text{const}} = 1$; $\sum_{j \in \mathcal{X}_s} \tau_{s;j} = 1$ for all $s \in V$; and $\sum_{j \in \mathcal{X}_s} \tau_{st;jk} = \tau_{t;k}$ for all $(s \to t) \in E$ and $k \in \{0, 1\}$. We specify a dual solution $\tau^*$ based on sets $\chi_s(\mathbb{S})$ and $\chi_{st}(\mathbb{S})$ for nodes $s \in V$ and edges $(s, t) \in E$ in the following way:

- for each node $s \in V$, we define the vector $\tau_s^*$ as follows: If $\chi_s(\mathbb{S}) = \{j\}$, then $\tau_{s;j}^* = 1$, whereas if $\chi_s(\mathbb{S}) = \{0, 1\}$, then $\tau_{s;j}^* = \tau_{s;1-j}^* = 0.5$.
- The vector $\tau_{st}^*$ for each edge $(s, t) \in E$ is defined as follows. (i) If $\chi_{st}(\mathbb{S}) = \{(j, k)\}$, then $\tau_{st;jk}^* = 1$; (ii) If $\chi_{st}(\mathbb{S}) = \{(j, 0), (j, 1)\}$, then $\tau_{st;j0}^* = \tau_{st;j1}^* = 0.5$. (iii) If $\chi_{st}(\mathbb{S}) = \{(0, k), (1, k)\}$, then $\tau_{st;0k}^* = \tau_{st;1k}^* = 0.5$; and (iv) If $\{(j, k), (1-j, 1-k)\} \subseteq \chi_{st}(\mathbb{S})$, then $\tau_{st;jk}^* = \tau_{st;1-j,1-k}^* = 0.5$.

Any component $\tau_\alpha^*$ not specified in these rules is set to zero. It is easy to see that each case is covered exactly once (assuming that sets $\chi_s(\mathbb{S})$ and $\chi_{st}(\mathbb{S})$ are non-empty). Furthermore, it can be verified that $\tau^* \in \text{LOCAL}(G)$, so that it is dual feasible. Now let us compute the value of this dual feasible point, which is given by $\langle \bar{\theta}, \tau^* \rangle$. Since $\bar{\theta}$ is a reparameterization of $\widehat{\theta}$ (i.e., $\bar{\theta} \cong \widehat{\theta}$), it follows that $\langle \bar{\theta}, \tau^* \rangle = \langle \widehat{\theta}, \tau^* \rangle$ (see [6]).

Consider some index $\alpha = (s; j) \in \mathcal{I}$. If $\tau_{s;j}^* > 0$, then our construction implies that $j \in \chi_s(\mathbb{S})$. By Lemma 1, this implies that $\widehat{\theta}_{s;j} = 0$. Similarly, if for index $\alpha = (st; jk) \in \mathcal{I}$ we have $\tau_{st;jk}^* > 0$, then $(j, k) \in \chi_{st}(\mathbb{S})$ and $\widehat{\theta}_{st;jk} = 0$. Therefore, $\widehat{\theta}_\alpha \tau_\alpha^* = 0$ for any $\alpha \in \mathcal{I} \setminus \{\text{const}\}$, so that we have $\langle \bar{\theta}, \tau^* \rangle = \langle \widehat{\theta}, \tau^* \rangle = \widehat{\theta}_{\text{const}} \cdot 1 = \Phi_\rho(\vec{\theta})$, meaning that the dual value $\langle \bar{\theta}, \tau^* \rangle$ is equal to the value $\Phi_\rho(\vec{\theta})$ of a primal feasible solution. By strong duality, the pair $(\vec{\theta}, \tau^*)$ must be primal-dual optimal. $\square$

**Remarks:** Among other consequences, Theorem 4 establishes a connection between our results and previous work on the roof duality approach to binary integer programs [4, 1]. It involves maximizing a lower bound on the energy function, for which the dual is known as the Rhys relaxation. Hammer et al. [4] showed that optimal fractional solutions of Rhys relaxation are weakly persistent, meaning the set of variables with integer values retain those same values in at least one optimal solution of the original minimization problem. This result suggests a close link between the TRW-LP and roof duality. Indeed, it is possible to show [7] that Rhys relaxation and the TRW-LP are essentially equivalent. With this additional equivalence, the connections can be summarized as follows. On one hand, weak persistency of the Rhys relaxation plus Theorem 4 imply Theorem 2; on the other hand, Theorems 4 and 2 in conjunction imply Rhys weak persistency.

## 5 Experimental results

Our theoretical results in the preceding section show that for binary problems, any variable uniquely specified by TRW is guaranteed to be correct (Theorem 2), and that for submodular problems, the TRW algo-rithm is guaranteed to specify all variables (Theorem 3). These results suggest that the TRW algorithm should still perform well on *near-submodular* energy functions–meaning relatively close to a submodular problem. Accordingly, this section is devoted to an experimental investigation of the percentage of variables correctly determined by TRW, when applied to energy functions in which the following three parameters are varied: the percentage $\alpha$ of submodular edges, the problem size $N$, and the strength of the singleton potentials ($\theta_s$) relative to that of the pairwise potentials ($\theta_{st}$). Note that this last parameter can be interpreted as a type of signal-to-noise ratio (SNR).

We provide experimental results on two types of graphs: $N \times N$ grids with 4-nearest neighbor interactions, and complete graphs on $N$ nodes. In all cases, we generated single-node potentials as Gaussians $\bar{\theta}_{s;0}, \bar{\theta}_{s;1} \sim \mathcal{N}(0,1)$, independently for each node. Pairwise potentials were set as $\bar{\theta}_{st;00} = \bar{\theta}_{st;00} = 0$ and $\bar{\theta}_{st;00} = \bar{\theta}_{st;00} = \lambda_{st}$, where the random variable $\lambda_{st} \sim |\mathcal{N}(0,\sigma^2)|$ with probability $\alpha$, and $\lambda_{st} \sim -|\mathcal{N}(0,\sigma^2)|$ with probability $(1-\alpha)$. We applied the sequential TRW algorithm [6] to each problem, terminating if the value of the lower bound (7) did not increase for 10 iterations. This criterion is appropriate since the sequential TRW updates are guaranteed [6] to never decrease the bound; moreover, if vector $\vec{\theta}$ does not satisfy WTA condition, then the bound is guaranteed to increase in a finite number of iterations. We used the condition $|\widehat{\theta}_{s;0} - \widehat{\theta}_{s;1}| > 10^{-6}$ to determine when $s \in V^{\text{fix}}$. For each triple $(\alpha, \sigma, N)$, we ran the algorithm on 100 sample problems and report the average percentage of correctly specified random variables. Note that there is no need to perform brute force exact computations to make this comparison, since any variable in $V^{\text{fix}}$ is guaranteed to be correctly specified (by Theorem 2).

Our first set of experiments examines the dependence of the correct fraction $p_{\text{cor}}$ on the potential strength $\sigma$ at a fixed value of $\alpha$. We fixed $\alpha = 0.5$ for grids and $\alpha = 0$ for the complete graphs; note these values give the worst behaviour for corresponding problems[1]. The results are shown in panels (a) and (b) of Figure 2, where we plot the correct percentage $p_{\text{cor}}$ on the vertical axis versus the edge strength $\sigma \cdot d$ along horizontal axis, where $d$ is the node degree. As expected, for small values of $\sigma$, the fraction $p_{\text{cor}}$ of correct variables is near 100%; moreover, as shown in panel (b) for complete graphs, the dependence remains approximately invariant as the graph size increases. In the second set of experiments, we fixed the size of the graph to $N = 32$, and measured the percentage of correct variables $p_{\text{cor}}$ as a function of $\alpha$ for different values of $\sigma \cdot d$. As shown in panels (c) and (d) of Figure 2, if the problem is near-submodular (i.e., $\alpha \approx 1$), then almost all variables are correctly specified ($p_{\text{cor}} \approx 100\%$).

# 6 Conclusion

This paper provides several theoretical guarantees on tree-reweighted (TRW) max-product algorithm as applied to problems with binary variables. We showed that TRW message-passing is always successful for submodular energy functions; moreover, for arbitrary energy functions, we proved that any fixed point that satisfies weak tree agreement can be used to specify a subset of a globally optimum solution. Experimental results show that for certain regimes of near-submodular functions, the TRW method continues to determine a relatively high fraction of the optimal solution. While the current paper focused on the binary case, the TRW method itself applies to arbitrary pairwise MRFs. Our current results in their precise form cannot be extended beyond the binary case (as there exist counterexamples); however, it should be possible to prove related results in the more general setting.

---

[1] The grid is invariant to the change $\alpha \leftarrow 1-\alpha$, which is equivalent to flipping all "odd" nodes of a bipartite graph.